\documentclass[twoside,leqno,twocolumn]{article}
\usepackage{ltexpprt}
\newcommand*\samethanks[1][\value{footnote}]{\footnotemark[#1]}

\usepackage{float}
\usepackage{amsmath}
\usepackage{graphicx}
\usepackage[show]{chato-notes}
\usepackage{epsfig}
\usepackage{url}
\usepackage{hyperref}
\usepackage{caption}
\usepackage{adjustbox}
\usepackage{booktabs}     
\usepackage{multirow}
\usepackage{siunitx}	
\usepackage{rotating}
\usepackage{enumerate}
\usepackage{algorithm}
\usepackage{algpseudocode}
\algnewcommand{\LineComment}[1]{\Statex \(\triangleright\) #1}
\algnewcommand{\LineCommentSpaced}[2]{\Statex \(#2 \triangleright\) #1}

\usepackage{wrapfig}

\usepackage{subcaption}

\usepackage[utf8]{inputenc} 
\usepackage[T1]{fontenc}    
\usepackage{url}            
\usepackage{booktabs}       
\usepackage{amsfonts}       
\usepackage{nicefrac}       
\usepackage{microtype}      

\newcolumntype{R}[2]{%
    >{\adjustbox{angle=#1,lap=\width-(#2)}\bgroup}%
    l%
    <{\egroup}%
}

 \newcommand*{\qedhere}{\hfill\ensuremath{\square}}%

\newcommand{\lrdiscovery}{\textsc{LRDiscovery}}

\newcommand{\bS}{\ensuremath{\mathbf{S}}}
\newcommand{\bT}{\ensuremath{\mathbf{T}}}

\newcommand{\bM}{\ensuremath{\mathbf{M}}}
\newcommand{\bX}{\ensuremath{\mathbf{X}}}

\newcommand{\bv}[1]{\ensuremath{\mathbf{v_{#1}}}}

\newcommand{\pimac}[2]{ \ensuremath{\pi{(\mathbf{#1,#2})}}}
\newcommand{\gamm}[2]{\ensuremath{\gamma{(\mathbf{#1,#2})}}}

\newcommand{\wholeM}{\bM}
\newcommand{\wholeMest}{\ensuremath{\hat{\bM}}}
\newcommand{\wholeMo}{\ensuremath{{\wholeM}_\Omega}}
\newcommand{\wholeMi}{\ensuremath{{\wholeM}_i}}
\newcommand{\subM}{\bS}

\newcommand{\subMi}{\ensuremath{{\subM}_i}}
\newcommand{\subMest}{\ensuremath{\hat{\subM}}}
\newcommand{\compM}{\bT}

\newcommand{\compMi}{\ensuremath{{\compM}_i}}
\newcommand{\compMest}{\ensuremath{\hat{\compM}}}
\newcommand{\rowSub}{\ensuremath{R_s}}
\newcommand{\colSub}{\ensuremath{C_s}}
\newcommand{\rowComp}{\ensuremath{R_t}}
\newcommand{\colComp}{\ensuremath{C_t}}
\newcommand{\singvecsub}[1]{\ensuremath{\mathbf{s_{#1}}}}
\newcommand{\singveccomp}[1]{\ensuremath{\mathbf{t_{#1}}}}
\newcommand{\singvecwhole}[1]{\ensuremath{\mathbf{v_{#1}}}}

\newcommand{\etal}{{et al.}}

\newcommand{\error}{{\ensuremath{\textsc{RelError}}}}

 \newcommand{\local}{targeted}

\newcommand{\lmafit}{{\tt LMaFit}}

\newcommand{\thealgo}{\texttt{SVP}}

\newcommand{\localmc}{{\tt Targeted}}
\newcommand{\partmc}{{\tt Partion}}
\newcommand{\targeted}{{\localmc}}

\newcommand{\spara}[1]{\smallskip\noindent{\bf{#1}}}

\begin{document}

\title{Targeted matrix completion\thanks{This research was supported in part by NSF grants CNS-1618207, IIS-1320542, IIS-1421759 and CAREER-1253393, as well as a gift from Microsoft.}}
\author{Natali Ruchansky\thanks{University of Southern California, ruchansk@usc.edu} \\
\and
Mark Crovella\thanks{Boston Unversity, \{crovella, evimaria\}@cs.bu.edu}
\and
Evimaria Terzi\samethanks}
\date{}

\maketitle


\begin{abstract} 

Matrix completion is a problem that arises in many data-analysis settings where the input consists of a partially-observed
matrix (e.g., recommender systems, traffic matrix analysis etc.). Classical approaches to matrix completion
assume that the input partially-observed matrix is low rank. The success of these methods depends on the
number of observed entries and the rank of the matrix; the larger the rank, the more entries need to 
be observed in order to accurately complete the matrix.
In this paper, we deal with matrices that are not necessarily low rank themselves, but rather they 
contain low-rank submatrices. We propose {\localmc}, which is 
a general framework for completing
such matrices. In this framework, we first extract the low-rank submatrices and then apply 
a matrix-completion algorithm to these low-rank submatrices as well as the remainder matrix separately. Although for the completion itself we use state-of-the-art completion
methods, our results demonstrate that {\localmc} achieves significantly smaller reconstruction errors than
other classical matrix-completion methods.  One of the key technical contributions 
of the paper lies in the identification of the low-rank submatrices from the input partially-observed matrices.
\end{abstract}

\section{Introduction}\label{sec:intro}

The problem of matrix completion continues to draw attention and innovation for its utility in data analysis.  Many datasets encountered today can be represented in matrix form; from user-item ratings in recommender systems, to protein interaction levels in biology, to source-destination traffic volumes in the city or Internet.  This type of measurement data is often highly incomplete, bringing about the need to estimate the missing entries which is precisely the goal of matrix completion.

Matrix data collected from many of these applications has also been observed to be low rank~\cite{davenportoverview,goldberg1992using,rennie2005fast}.
The assumption that the underlying matrix is low-rank is key to many matrix completion methods used today~\cite{cai2010singular,chen14coherent,foygel2011concentration,keshavan2009matrix,meka2009matrix,ruchansky2015matrix,wen2012solving}.
In fact, 
there is a direct relationship between the number of observed entries and the rank of the underlying matrix; the larger the rank, the more entries are required for an accurate
reconstruction to be produced. 

In this paper, we deviate from the strict low-rank assumption of the overall matrix
and we assume that the matrix has many lower-rank submatrices.
For example, in user-preference data, such submatrices may correspond to 
subsets of users that behave similarly with respect to only subsets of products. Not only is our assumption observed in practical settings, but is also the standard assumption in many recommender- system principles; for example, in collaborative filtering, recommendations are made to a user based on the preferences of other similar users~\cite{aggrawal16recommender,herlocker04evaluating,sarwar01itembased}. 
Further, the assumption is a common in the analysis of traffic networks, datasets in the natural sciences (e.g., gene expression data), and more. Despite the fact that the assumption of the existence of low-rank submatrices is prevalent to many data-analysis settings, there is very little work in the matrix-completion literature that exploits this structure; existing matrix-completion methods need to be enhanced to work well for matrices that contain low-rank submatrices.

The central contribution of our paper is that we  propose a new matrix completion paradigm
that allows for the completion 
of partially-observed matrices that contain low-rank submatrices.
This framework, which we call {\localmc}, works as follows: first, in a completely unsupervised
manner, it identifies partially-observed submatrices that we expect to be low rank. Then, using standard
matrix-completion s it completes each submatrix as well as the remainder of the matrix separately (once
the submatrices are removed).

Since most real-world datasets do not come with auxiliary information on the location of low-rank submatrices, we also study the problem of finding low-rank submatrices within a partially-observed matrix.  Therefore, a second technical contribution of our paper is the design of {\thealgo} (Singular Vector Projection), which
identifies low-rank submatrices in an unsupervised manner. Inspired by the Singular Value Decomposition, {\thealgo} is simple and efficient -- requiring only the computation
of the first left and right singular vectors of the input.

From the practical point of view, our experiments with generated data using different models demonstrate 
that {\thealgo} is extremely accurate in identifying low-rank submatrices in partially-observed data. With {\thealgo} as an important tool at hand, we demonstrate in the experiments that {\localmc} is able to significantly improve the accuracy of matrix completion both in real and synthetic datasets.

\section{Related work}\label{sec:related}

In this section, we highlight the existing work related to matrix completion, as well as the problem of finding a low-rank submatrix, which we call {\lrdiscovery}.

\spara{Matrix completion:}
Existing approaches to low-rank matrix completion span a wide range of techniques, from norm minimization~\cite{candes12exact}, to singular value thresholding~\cite{cai2010singular}, to alternating minimization~\cite{wen2012solving}, to name a few.  
What these approaches have in common is that they assume the whole matrix has low rank, and pose an optimization to fit a single rank-$r$ model to the entire matrix.  These algorithmss are different from ours since
they do not take advantage of the presence of low-rank submatrices.

To the best of our knowledge, the only matrix-completion algorithm that tries to exploit
the existence of multiple low-rank submatrices is an algorithm called {\tt LLORMA}, which was proposed by Lee {\etal}~\cite{lee2013local}.  
Although Lee {\etal} pose their task as a matrix factorization problem, at a high-level, they also argue that using several smaller factorizations as opposed to a single (global) factorization is more accurate and efficient.  The
{\tt LLORMA} algorithm samples a fixed number of submatrices using information about the distances between the
input rows and columns.  The output is a linear combination of the factorizations of the selected submatrices.
The main
difference between our framework and {\tt LLORMA} is that {\targeted} does not require 
knowledge of the distances between the rows and the columns of the matrix. 
Hence, {\targeted} is more efficient and practical.  
Another difference is that the {\tt LLORMA} algorithm relies on a good sampling of represen- tative submatrices and an adequate number of them, whereas the Targeted algorithm does not rely on such sampling. Further, these submatrices selected by {\tt LLORMA} are not necessarily low rank, since this is not the goal of the algorithm nor of the specific problem it addresses.

\spara{Low-rank submatrix discovery:}
Although there exists work on discovering structure in data, there is little work on the specific problem of discovering
low-rank submatrices embedded in larger matrices.
We review two lines of research that are most related to the {\lrdiscovery} problem. \\

To the best of our knowledge, Rangan~\cite{rangan2012simple} was the first to explicitly ask the question
of finding a low-rank submatrix from a larger matrix.  The work
focuses on a particular instance where the entries of the matrix \emph{and} the submatrix adhere to a standard Gaussian distribution with mean zero and variance one, and the submatrix has rank less than five.  
The algorithm, which we call {\tt BinaryLoops},
searches for a low-rank submatrix by comparing the $\pm$-sign patterns of the entries in rows and columns. 
The output is a nested collection of rows and a nested collection of columns, hence requiring an additional search on the collections to select the right set of rows and columns.
The underlying assumptions restrict the success of
{\tt BinaryLoops} to instances where the submatrix is large and has rank less than five.
Since our analysis does not make the same assumptions and uses the data values themselves, the {\thealgo} algorithm
succeeds on a larger range of submatrix sizes and ranks.
Furthermore, {\thealgo} directly outputs a subset of rows and columns indexing the discovered submatrix and does not require setting a tuning parameter.

The second line of related work,
is \emph{subspace clustering} (SC) as studied in~\cite{elhamifar2009sparse,vidal2014low}, though it does not address exactly the same problem.
SC assumes that the rows of the matrix are drawn from a union of \emph{independent} subspaces, ands
seeks to find a clustering of the rows into separate subspaces.
SC is related to {\lrdiscovery} in the special case where the low-rank submatrix is only a subset of the rows of the matrix, spanning all columns, and where the submatrix is also an independent subspace.   
In our experiments, we observe that these algorithms (appropriately modified for our problem)
cannot accurately separate the submatrix when the values in the submatrix are not significantly larger than the rest. Further, the algorithms are sensitive to the presence of large values elsewhere in the data.
In contrast, our analysis does not assume that the submatrix is an independent subspace and {\thealgo} proves to be much more resilient.

\spara{Planted Clique:}
Interestingly, the {\thealgo} algorithm we propose for finding a low-rank submatrix bears resemblances to the algorithm for the Planted Clique problem presented by Alon {\etal} in~\cite{alon1998finding}; both algorithms use some of the singular vectors of a matrix to discover a submatrix with a particular property.  Although a clique corresponds a rank-one submatrix of the adjacency matrix, a low-rank submatrix is not necessarily a clique.  Further, the success of the algorithm developed in~\cite{alon1998finding} relies on assumptions on the node degrees that allow bounds to be placed on the gap between the second and third eigenvalues.  Not only is it unclear that such assumptions would hold in our setting, but it is also not clear whether they are meaningful in the context of low-rank submatrices.

\section{Preliminaries}

Throughout we use {\wholeM} to refer to an $n\times m$ fully-known matrix $\wholeM\in\mathbb{R}^{n\times m}$ of rank $r$, with {\wholeMi} to denoting the $i$-th row of {\wholeM} and $\wholeM(i,j)$ the entry $(i,j)$. 
If {\wholeM}
is not completely known we use
$\Omega\subseteq \{(i,j)|1\leq i\leq n, 1\leq j\leq m\}$ to denote the subset of entries that are observed.  The partially observed matrix is denoted as {\wholeMo}, and when a matrix completion algorithm is applied to {\wholeMo} we use {\wholeMest} to denote the output estimate.
To measure the error of {\wholeMest} we use the relative Frobenius error:
\begin{align}
{\tt RelErr}(\wholeMest)=\frac{\|\wholeM-\wholeMest\|_f^2}{ \|\wholeM\|_f^2}\label{eq:error}
\end{align}

For a low-rank submatrix {\subM} of {\wholeM}, we use $\rowSub\subseteq \{1,\ldots , n\}$ and $\colSub\subseteq \{1,\ldots ,m\}$
to denote the rows and columns of the original matrix that fully define {\subM}. That is, $\subM = \wholeM(\rowSub,\colSub)$.
For a set of rows $\rowSub$ we use the term \emph{complement} to refer to the set of rows
$\overline{\rowSub}=\{\{1,\ldots ,n\}\setminus \rowSub\}$. 
We also use $\compM$ to denote the \emph{complement} of
$\subM$, that is if $\overline{\rowSub} = \{1,\ldots , n\}\setminus \rowSub$ and  $\overline{\colSub} = \{1,\ldots , m\}\setminus \colSub$,
then $\compM = \wholeM(\overline{\rowSub},\overline{\colSub})$.  

The notation {\bv 1} is reserved for the first singular vector of {\wholeM}, and for any other (sub)matrix $\subM$ we use the corresponding lowercase {\singvecsub i} to denote the $i$-th right singular vector of {\subM}.

Finally, we use the following conventions: $\|\cdot\|$ as shorthand for the $L_2$ norm $\|\cdot\|_2$ of a vector or matrix.
Recall, that the $L_2$ norm of a matrix is its first singular value.
For vectors $\mathbf{x}$ and $\mathbf{y}$,
  we also use  $\langle \mathbf{x},\mathbf{y} \rangle$ to denote the inner product of
$\mathbf{x}$ and $\mathbf{y}$.\label{sec:notation}

\section{The  {\localmc} matrix-completion framework}\label{sec:targeted}

The {\targeted} framework for matrix completion takes as input a partially-observed matrix $\wholeMo$
and proceeds in three main steps: first, identify low-rank submatrices, then separate them from the rest of the data,
and finally, complete the extracted low-rank submatrices as well as the whole matrix using 
an existing matrix completion algorithm. 
The steps of {\targeted} are described in more detail below as well as in upcoming sections.

\spara{Step 1) Identification of low-rank submatrices:}  In this step, {\targeted} deploys
our own low-rank submatrix identification technique, which we call {\thealgo} and we 
describe in full detail in Section~\ref{sec:lrsd}.

\spara{Step 2) Separation of the low-rank submatrices:}
In this step, {\targeted} separates the low-rank submatrices from the rest of the data.  
To do this we extract each submatrix {\subM}, and then replace the entries of {\subM} with zeros, i.e. $\wholeMo(\rowSub, \colSub)=0$ for each {\subM}.

\spara{Step 3) Matrix completion:} For the completion phase we deploy
a state-of-the-art matrix completion algorithm
in order to complete all the extracted low-rank submatrices, found in Step 1, as well as the remaining
matrix, constructed in Step 2. For our experiments, we use the {\tt LMaFit}~\cite{wen2012solving} 
matrix completion algorithm
because it is extremely accurate and efficient in practice.

To the best of our knowledge, Step 1 has not been fully addressed in the literature to date.  In fact, the problem of finding a low-rank submatrix in a fully known, let alone partially known, matrix is an open and interesting problem.  Hence before investigating the benefit of a {\targeted} framework for matrix completion, we 
describe our algorithm for Step 1 in the next section.

\section{Low-rank submatrix discovery}\label{sec:lrsd}
In this section, we study the first step of {\targeted}: the problem of finding a low-rank submatrix which we call {\lrdiscovery}.   
We first focus on developing an algorithm for the setting where the matrix is fully known, and provide an analysis of when it is likely to succeed.  Next, we extend to the partially-observed setting.

\subsection{Exposing low-rank submatrices.}\label{sec:lowranksubm}


First, we lay out our main analytical contribution in which we characterize when the singular vectors of {\wholeM} can be used to expose evidence that there exists a low-rank submatrix {\subM} in {\wholeM}.
We focus the discussion on 
finding the rows of this submatrix (i.e., we fix ${\colSub}=\{1\dots m\}$), but the analysis is symmetric for the columns. 
The following two parameters are key to our 
characterization:
\[{\pimac S T}=\frac{\|{\subM}\|^2}{ \|{\compM}\|^2}\text{ and  }  {\gamm S T}= \max_{j}{\langle {\singvecsub 1},{\singveccomp j}\rangle}\]

The first parameter measures the magnitude of the $L_2$-norm of {\subM} with respect to {\compM}.
The second parameter, {\gamm S T},  measures the geometric orientation of {\subM} with respect to {\compM}. Intuitively, when $\gamma$ is small,  {\subM} and {\compM} are well-separated. When 
{\subM} and {\compM} are clear from the context we use $\pi$ to denote ${\pimac S T}$ and $\gamma$ to denote {\gamm S T}.

The two parameters allow us to characterize the behavior of {\singvecwhole 1} with respect to the magnitude 
and geometry of {\subM} in {\wholeM}.  In particular, we show that when $\pi$ is large and $\gamma$ is 
adequately small with respect to $\pi$, the normalized projection of a row {\subMi} on {\singvecwhole 1} will be \emph{larger} 
than that of {\compMi} on {\singvecwhole 1}. 
The gap between the projections suggests an approach to {\lrdiscovery} that uses {\singvecwhole 1} to expose evidence of {\subM}.
To simplify the discussion and capture the intuition above we define:
 \[\Delta_{\subM,\compM}=\frac{1}{ |\rowSub|}\|\mathcal{N}(\subM) {\singvecwhole 1}\|_1- \frac{1 }{|\rowComp|}\|\mathcal{N}(\compM){\singvecwhole 1} \|_1\]
 Here $\mathcal{N}(\bX)$ is an operator that normalizes each row of matrix $\bX$.   Hence, the above $\Delta$ is designed to capture the differences in the average projection of rows of {\subM} and {\compM} on the first singular vector {\singvecwhole 1} of {\wholeM}. 

\noindent
The crux of our analysis lies in the following proposition:\begin{proposition}
If $\pi>1$ and $\gamma<\epsilon(\pi-1)$, then $\Delta_{\subM,\compM}> (1-\gamma)(1-\epsilon)$, i.e. $\Delta_{\subM,\compM}$ is $\epsilon$-close to $(1-\gamma)$.\label{prop:sep}
\end{proposition}
Intuitively, Proposition~\ref{prop:sep} indicates that if {\subM} has 
large $L_2$-norm and is geometrically well-separated from {\compM}, then the normalized projection of {\wholeMi} on {\singvecwhole 1} will be larger for $i\in \rowSub$ than for $i\notin \rowSub$ -- larger by approximately $(1-\gamma)$.   This separation suggests that the projections can be used as a feature for separating {\subM} from {\compM}.

\emph{Proof sketch of Proposition~\ref{prop:sep}.}   Due to space limitations, we sketch the proof of Proposition~\ref{prop:sep} through the case in which {\wholeM}  consists of 
two rank-one submatrices: $\subM=\wholeM\left(\rowSub,:\right)$ and its complement $\compM=\subM\left(\overline{\rowSub},:\right)$.   Recall that the first singular vector of {\subM} is {\singvecsub 1} and the first singular vector of {\compM} is {\singveccomp 1}.

A key first step is to show that {\singvecwhole 1} can always be expressed as a (linear) combination of just {\singvecsub 1} and {\singveccomp 1}, i.e. ${\singvecwhole 1}=\alpha{\singvecsub 1}+\beta{\singveccomp 1}$.  We derive expressions for $\alpha$ and $\beta$, showing that
\[ \alpha= c\sigma \big({\singvecsub 1}(n)  \big( \lambda-\tau^2\big) -{\singveccomp 1}(n)\big(\sigma\tau\gamma \big) \big)\]  \[
\beta= c\tau \big({\singveccomp 1}(n)  \big( \lambda-\sigma^2\big) -{\singvecsub 1}(n)\big(\sigma\tau \gamma \big) \big) 
\]
in which $\sigma=\|\subM\|,$ $\tau=\|\compM\|$, $\lambda=\|\wholeM\|^2$, ${\singvecsub 1}(n)$ is the $n$-th entry of {\singvecsub 1}, and $c$ is a constant. 
Next we show that $\lambda$ can be expressed in terms of $\sigma$, $\tau$, and $\gamma$:
\[\lambda=\frac{1}{2}\big( \sigma^2+\tau^2\pm\sqrt{((\sigma^2-\tau^2)^2+4\sigma^2\tau^2\gamma^2)} \big)\]
Putting together the above, it is possible to obtain expressions for $\langle {\singvecwhole 1}, {\singvecsub 1}\rangle$ and $\langle {\singvecwhole 1}, {\singveccomp 1}\rangle$ in terms of $\pi$ and $\gamma$.  Finally, $\Delta_{\subM,\compM}$ can be expressed in terms of these inner products and simplifying we obtain Proposition~\ref{prop:sep}.\qedhere

\spara{Toy case of rank-1:}
To explore the implications of Proposition~\ref{prop:sep}, we continue to examine the case in which {\wholeM}  consists of 
two rank-one components.   
In this case, $\gamma={\gamm \subM \compM}= |\langle {\singvecsub 1},{\singveccomp 1} \rangle|$.
We claim that the set of $|\langle {\singvecwhole 1},\mathcal{N}({\wholeMi}) \rangle|$ are either identical for all {\wholeMi}, or take one of two values:
\begin{equation}\nonumber
|\langle {\singvecwhole 1},\mathcal{N}({\wholeMi})  \rangle|=
\begin{cases}
| \langle {\singvecwhole 1}, {\singvecsub 1}\rangle | , & \text{if}\  i\in \rowSub \\
| \langle {\singvecwhole 1}, {\singveccomp 1}\rangle |, & \text{otherwise}
\end{cases}
\end{equation}
Since {\subM} and {\compM} are rank one, $\mathcal{N}({\wholeMi}) =\frac{{\wholeMi}}{ \|{\wholeMi}\|}={\singvecsub 1} \, \forall i\in \rowSub$ and $\mathcal{N}({\wholeMi}) ={\singveccomp 1} \, \forall i \in \overline{\rowSub}$.  Thus, the gap between the projections of {\subM} and the projections of {\compM} is
  $\Delta_{\subM,\compM}=|\langle {\singvecwhole 1}, {\singvecsub 1} \rangle | - | \langle {\singvecwhole 1}, {\singveccomp 1}\rangle|$.  Using Proposition~\ref{prop:sep}, we can say that if the specified conditions on $\pi$ and $\gamma$ hold, then $\Delta_{\subM,\compM}$ is $\epsilon$-close to $(1-\gamma)=1-|\langle {\singvecsub 1},{\singveccomp 1}\rangle |$.   Hence, as long as {\singvecsub 1} and {\singveccomp 1} are not parallel, the gap $\Delta_{\subM,\compM}$ between {\subM} and {\compM} with respect to {\singvecwhole 1} is nonzero, and can be used to find the indices of {\subM}.

\spara{The case with larger ranks:}
In the case where {\subM} and its complement {\compM} have arbitrary ranks, the  
situation is not so simple.  The complication comes from the fact even if $i\in \rowSub$ the corresponding projection on {\singvecwhole 1} may not be exactly $|\langle {\singvecwhole 1},{\singvecsub 1}\rangle |$, and in particular it may be small. 
The generalization of Proposition~\ref{prop:sep} to larger ranks can be made using the claim that there is still a subset $R_{s'}\subseteq \rowSub$ that will have a large projection on {\singvecwhole 1}.\begin{proposition} If the rank-$\ell$ submatrix {\subM} is nondegenerate, $\pi>1$, and 
$\gamma=\epsilon(1-\pi)$, then 
$\exists R_{s'}\subset \rowSub$ with $|R_{s'}|\geq \ell$ such that 
$\Delta_{{\subM }',{\compM}}$ is $\epsilon$-close to $ 1-\gamma$.
  \label{prop:largerank}
  \end{proposition}
 An $n\times m$ rank $r$ matrix is \emph{nondegenerate} 
 if there is no singular submatrix smaller than $(r+1)\times (r+1)$.

The intuition of the proof of
 Proposition~\ref{prop:largerank} follows that of the rank-1 case.
 The first step is to derive an expression for {\singvecwhole 1} in terms of the singular vectors {\singvecsub i} and {\singvecsub j}, and expressions for the subsequent inner products with {\singvecwhole 1}.  
 Next, we consider $\Delta_{\subM, \compM}$, and using nondegeneracy we claim that there is a subset $R_{s'}$ of the rows of {\subM} that will have high inner product with {\singvecsub 1}.  We can then show that for each {\singveccomp i}, $\Delta_{{\singvecsub 1},{\singveccomp i}}=\langle {\singvecwhole 1}, {\singvecsub 1}\rangle - \langle {\singvecwhole 1},{\singveccomp i} \rangle$ is $\epsilon$-close to $1-\langle {\singvecsub 1}, {\singveccomp i}\rangle$.  Finally, putting everything together we can show that $\Delta_{\subM',\compM}$ is $\epsilon$-close to ($1-\gamma$).

 In the above discussion we focused on subsets of rows (i.e. the case where $\subM = \wholeM(\rowSub,:)$).  The results
 generalize to the case where {\subM} is a subset of both rows and columns ($\subM = \wholeM(\rowSub,{\colSub})$), by applying the above analysis recursively on $\wholeM(\rowSub,:)$ and $\wholeM(:,{\colSub})$.  Hence, we have expressed conditions under which low-rank submatrices can be discovered by the first singular vectors of a matrix, namely: $\pi>1$ and $\gamma<\epsilon(1-\pi)$.    
 As we will demonstrate experimentally in Section~\ref{sec:experiments}, these conditions are natural in real world data.  An intuitive example is a group of users $U$ whose rating patterns for a subset of movies $V$ are highly similar, yet differ from other users and on other movies.  
 The conditions lead to an intuitive algorithm that succeeds on a broader set of problem instances than any existing related approach.

\subsection{Extracting low-rank submatrices.}\label{sec:algo}

In this section, we provide a simple and effective algorithm for finding a low-rank submatrix {\subM}.
Our algorithm builds upon the results of the previous section, which imply the following:
the projections of {\wholeMi} on ${\singvecwhole 1}$ 
for $i\in \rowSub$ will have larger values than the corresponding projections of {\wholeMi} for $i\notin \rowSub$.  
Hence, a straightforward way to find {\subM} is to project {\wholeM} onto {\singvecwhole 1} and partition the projections into high and low values. This is exactly the idea behind our algorithm for finding {\subM}.  

\newpage\noindent
The pseudo code of this algorithm,
which we call {\thealgo} standing for Singular Vector Projection,  is below:
\begin{enumerate}
	\item {\singvecwhole 1}=${\tt SVD}(\wholeM)$ \\
	Compute the first right singular vector of {\wholeM}.
	\item $\mathbf{p}=$\textbf{Project}$(\wholeM,{\singvecwhole 1})$. \\ 
	Construct the vector $\mathbf{p}=\{p_1,\dots , p_n\}$ of projections on the singular vector $p_i=\langle {\singvecwhole 1}, \frac{\wholeMi}{ \|\wholeMi\|}\rangle$.
	\item $\{\rowSub,\overline{\rowSub}\}=\text{\textbf{Partition}}(p)$ \\
	Partition the rows of {\wholeM} into two groups based on the corresponding values in $\mathbf{p}$. 
	Identify the set of rows $\rowSub$ to be the group that has the largest average of the corresponding $p_i$ values.
	\item \textbf{Repeat} steps (1) and (2) on $\wholeM^T$ to find $\colSub$.
\end{enumerate}
The {\thealgo} algorithm consists of two main steps (applied independently to the rows and the columns of {\wholeM}).
After computing {\singvecwhole 1}, the rows (resp.\ columns) of {\wholeM} are normalized and projected on ${\singvecwhole 1}$. These $1$-dimensional points are then clustered into two clusters -- in our implementation we use 
$k$-means.  The cluster of rows (resp.\ columns) with the largest mean is
assigned to {\subM}.
Hence, following Proposition~\ref{prop:sep}, we can intuitively say that \emph{if $\pi$ is large and $\gamma$ is small then the low-rank submatrix in {\wholeM} is discoverable by {\thealgo}.}

A reasonable extension is to create multi-dimensional $\mathbf{p}_i$'s by projecting onto more than one of the singular vectors of {\wholeM}.  Our experiments indicated that on average, using three singular vectors improves the accuracy with a minimal effect on the efficiency; using more than three singular vectors did give any significant improvement, and we conjecture that this is tied to our use of k-means.

To find multiple low-rank submatrices, {\thealgo} can be run on the matrix formed by removing the rows and columns of {\subM}, i.e. $\wholeM(\rowComp,\colComp)$.  If the number of low-rank submatrices in {\wholeM} is known a priori, {\thealgo} can be repeated exactly this many times.  Otherwise, the difference in the average projection between the output partitions, $\Delta_{\hat{\subM},\hat{\bT}}$, can be used as a stopping criterion where $\hat{\subM}$ is the output of {\thealgo}.  For example, allow the algorithm to keep iterating while $\Delta_{\hat{\subM},\hat{\bT}}$ is greater than a threshold, and stop when it falls below; in our experiments, we found $0.2$ to work well.
 Another possibility is to replace the submatrix with random noise that is small relative to the values in the rest of the data and rerun on the modified matrix; such an approach would allow for the discovery of submatrices that overlap in rows or columns.

\subsection{Dealing with incomplete matrices.}\label{sec:incomplete}

The main tool that {\thealgo} relies on is finding the first singular vectors of a matrix.  Since the Singular Value Decomposition of an incomplete matrix is not defined, extending {\thealgo} to handle incomplete data translates to developing an approach to estimating the singular vectors of a matrix with missing entries.  
To do this we follow the Incremental-SVD approach first described by Simon Funk in the context of the Netflix challenge~\cite{funksvd}, and later expanded and implemented in the IncPack toolbox~\cite{incpack}.   Hence Step 1 in the {\thealgo} described in Section~\ref{sec:algo} is replaced with: {\singvecwhole 1}={\tt Incremental-SVD}$(\wholeMo)$.

We experimented with other approaches such as treating the missing entries as zeros and applying the regular {\tt SVD} algorithm.  Further, since this approach may result in over-fitting we modified an implementation of a QR-based power method to stop at an early iteration when the value of $\Delta_{\subMest, \compMest}$ stabilizes.  
All three of these methods gave similar results, without a clear and consistent winner. Finally, we considered an aggregate of the three approaches using clustering aggregation techniques (such as by Gionis {\etal} in~\cite{gionis2007clustering}) which slightly improved accuracy at the cost of efficiency.

We note that in principle we require accurate estimation of the direction of the singular vector, as opposed to accurate estimation of the matrix entries; however, we are not aware of any approach for this goal.

\section{Experiments}\label{sec:experiments}

In this section we answer the question posed in Section~\ref{sec:intro}, namely: \emph{Can the knowledge of the existence of a low-rank submatrix improve the accuracy of completion?}  We do this by evaluating the performance of  {\localmc}   on datasets with missing entries, and demonstrate the large improvement in accuracy over the state-of-the-art matrix completion algorithms.

For our comparison, we use {\tt LMaFit}~\cite{wen2012solving} as 
a representative matrix-completion algorithm
that operates on the whole matrix, assuming it is low-rank.  Again,
the reason we picked {\tt LMaFit} is because it is extremely accurate and efficient in practice.
In our experiments, 
 we consider both synthetic and real world data.
After this main set of experiments, we study the behavior of {\thealgo} (step 1 in the {\localmc} framework) over a wide range of matrix instances, and show that {\thealgo} succeeds in more cases than the comparison algorithms.

\spara{Setup:}
Throughout the experiments we generate a $1000\times 1000$ random matrix {\wholeM} of rank $r$ with entries following a standard Gaussian distribution with mean equal to zero and variance equal to one, we call this the \emph{background matrix}.  We then plant a $100\times 100$ submatrix {\subM} of rank $\ell$ with entries adhering to a centered Gaussian distribution.  
We maintain $\pi$ constant at $\pi=1.2$; as this value is only
slightly larger than 1, the instances we generate are hard for {\thealgo}.

\subsection{Evaluation of {\local} matrix-completion}

To quantify the improvement in matrix completion when using {\localmc}, we compare the accuracy of the completions produced by each approach.
We setup the experiments as described above and set the rank of the matrix {\wholeM} to a low rank $r=30$ and the rank of the submatrix {\subM} to a lower rank $\ell=2$. 
To form the partially observed {\wholeMo} we vary the percentage of observed entries from $0\%$ to $100\%$, and measure the relative error ({\error}) of each completion according to Equation~\ref{eq:error} in Section~\ref{sec:notation}.

\begin{figure}[H]
\centering
\includegraphics[scale=.5]{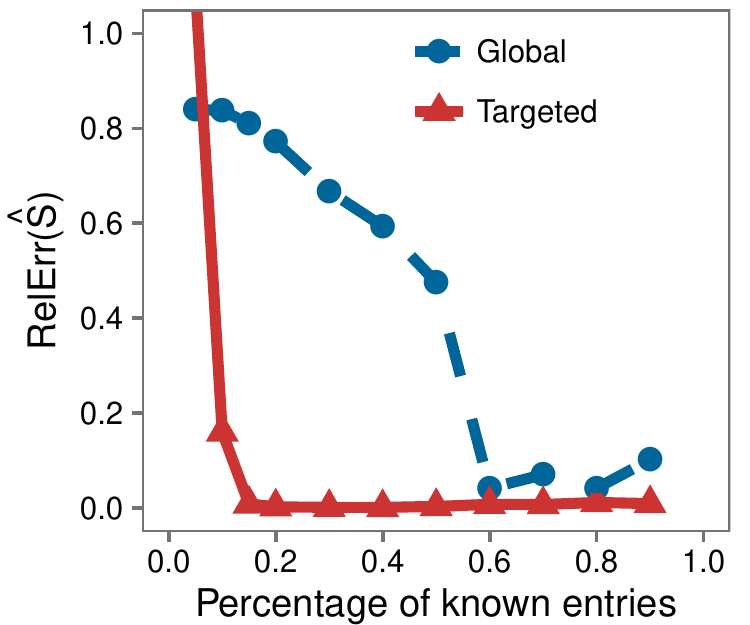}
\includegraphics[scale=.5]{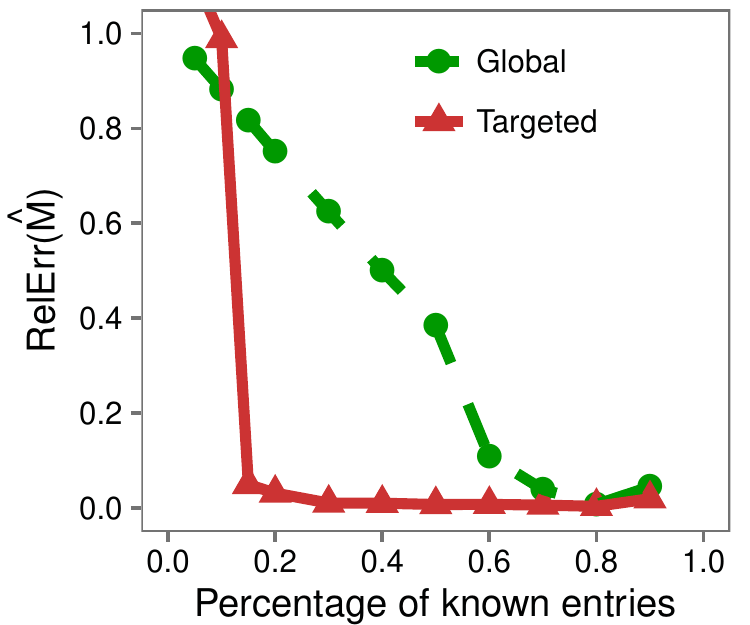}
\caption{~\label{fig:comp} {\error} of completion on synthetic data.}
\end{figure}

 Figure~\ref{fig:comp} shows the relative error over {\subMest} (left) and over {\wholeMest} (right).  Immediately we observe the large increase in accuracy when using the {\localmc} framework.  Whereas {\lmafit} requires approximately $60\%$ of the entries to be known to achieve a relative error lower than $0.2$, {\localmc} completion only needs about $20\%$ of the entries for the same accuracy.

\begin{figure*}[h]
\centering
	\captionsetup[subfigure]{justification=centering}
\begin{subfigure}{0.24\textwidth}
	\centering
	\includegraphics[scale=.5]{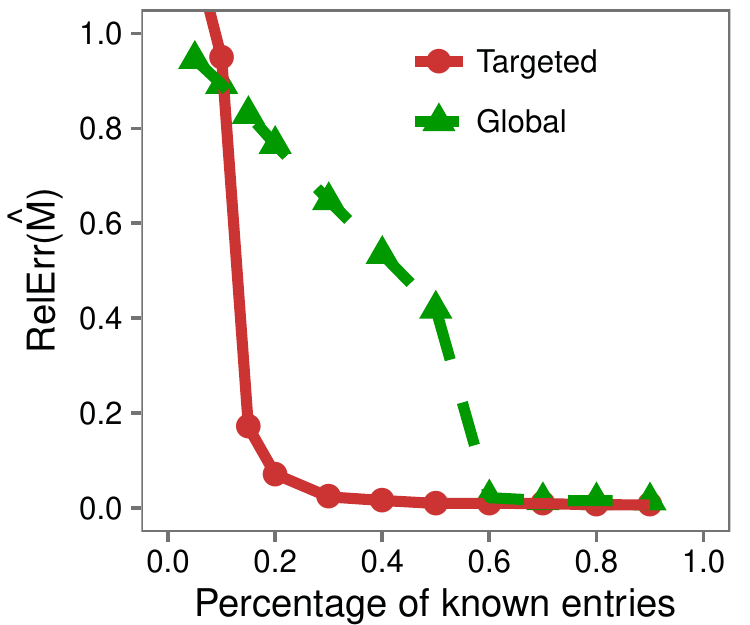}
\caption{Error over {\wholeM}} \label{fig:synthM}
\end{subfigure}
\begin{subfigure}{0.24\textwidth}
	\centering
	\includegraphics[scale=.5]{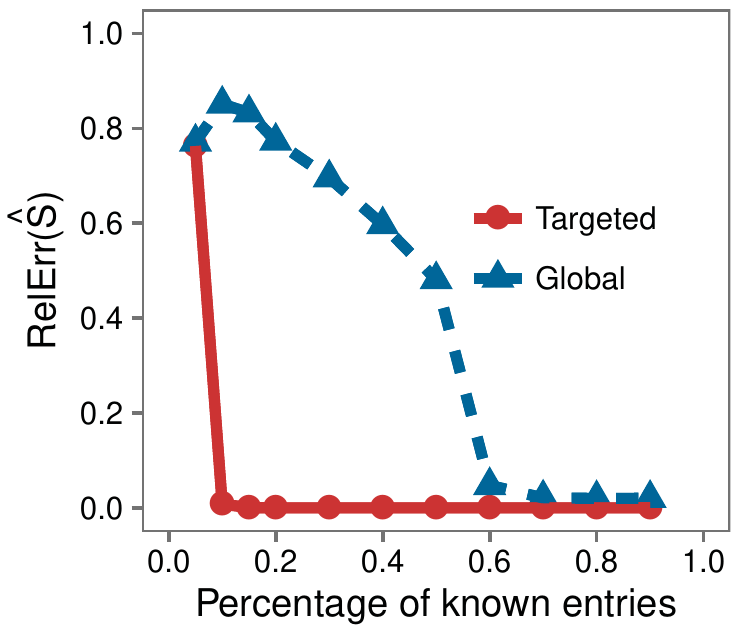}
\caption{Error over first {\subM}} \label{fig:synthS1}
\end{subfigure}
\begin{subfigure}{0.24\textwidth}
	\centering
		\includegraphics[scale=.5]{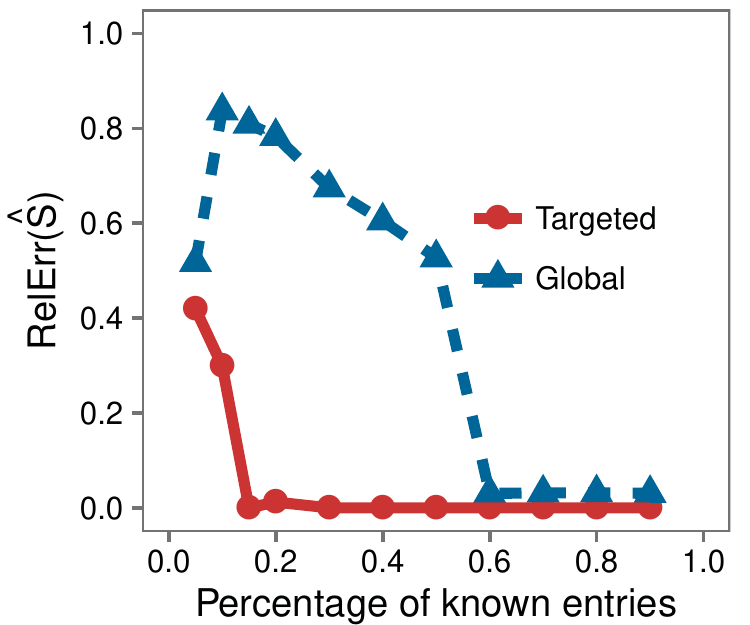}
\caption{Error over second {\subM}} \label{fig:synthS2}
\end{subfigure}
\begin{subfigure}{0.24\textwidth}
	\centering
		\includegraphics[scale=.5]{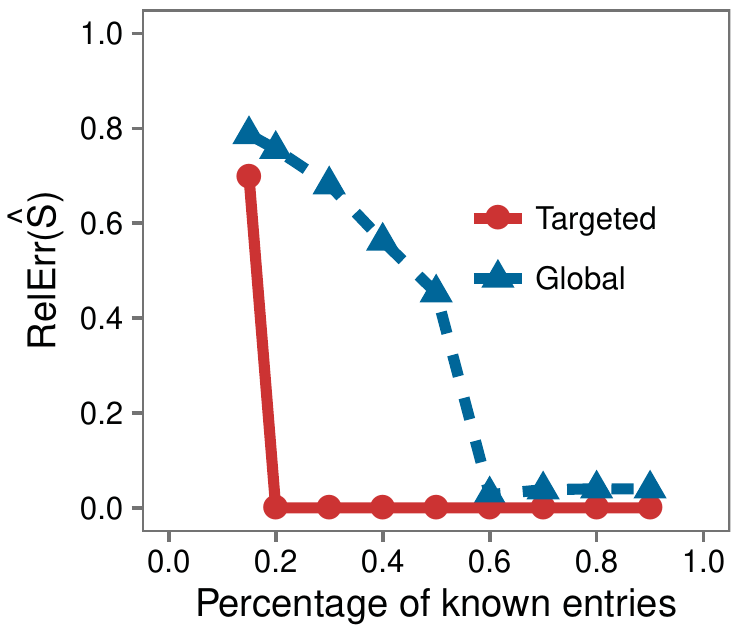}
\caption{Error over third {\subM}.} \label{fig:synthS3}
\end{subfigure}
\caption{\label{fig:compmult} Relative error in completion on a matrix with multiple low-rank submatrices.}
\end{figure*}

To test the accuracy when the matrix contains multiple low-rank submatrices, we follow the same setup but plant three low-rank submatrices of equal size and rank.
In Figure~\ref{fig:compmult}, we see that the results are consistent for each submatrix; we observe that the {\localmc} framework achieves a large improvement over {\lmafit} for each submatrix, and for the whole matrix.

One insight into the success of {\localmc} is that a single model is not enough for standard matrix completion to capture a matrix with low-rank submatrices.  Hence, a {\localmc} framework that isolates the low-rank submatrices and completes them separately is much more accurate.
An observation we made in practice was that the success of {\localmc} was also due to the reliance of matrix completion on accurate rank-estimation.  Techniques for estimating the rank of a partially-observed matrix often err when the input contains low-rank submatrices.  However, when applied separately to each component, both the rank estimate and consequently the completion are more accurate.

\spara{Sanity check:}  
We wish to check whether the benefit of {\localmc} can be attributed to the fact that the completion algorithm is applied to submatrices.  We do this by asking: \emph{given a matrix {\wholeM} that does not have low-rank submatrices, will the accuracy of completion increase if the algorithm is run on submatrices?}
\begin{wrapfigure}{r}{0.23\textwidth}
	\centering
\includegraphics[scale=.5]{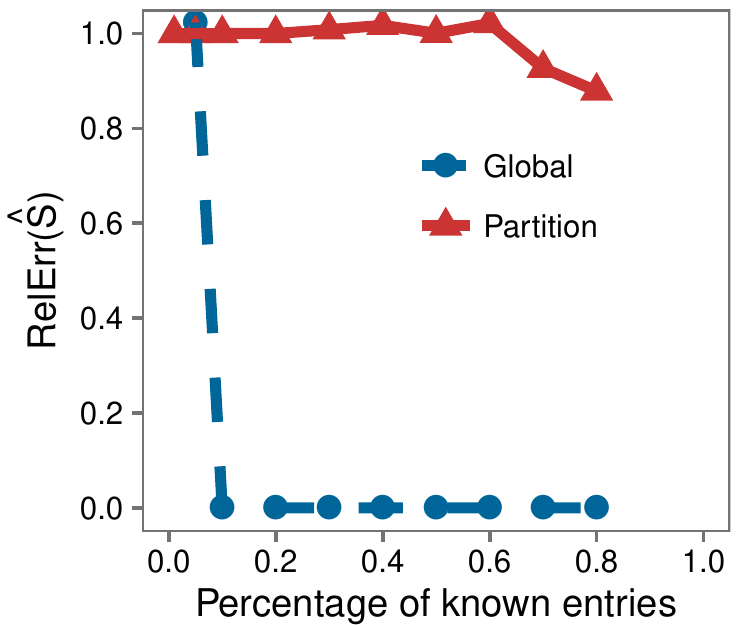}
\caption{~\label{fig:sanity} {\error} of {\partmc} in sanity check.}
\vspace{-1ex}
\end{wrapfigure}
We follow the same setup as Figure~\ref{fig:comp}, but on a matrix {\wholeM} without low-rank submatrices.  We selected a random $50\times 50$ submatrix {\subM} of {\wholeM}, and completed it separately to mimic {\localmc}; call this approach {\partmc}.
Figure~\ref{fig:sanity} shows the {\error} over {\subM} as a function of the percentage of known entries. We see that completing submatrices separately does not help the completion algorithm, implying that the accuracy of {\localmc} is not due solely to its application on smaller parts of the matrix.

\spara{Running time evaluation:}
To evaluate the change in running time when adopting {\localmc}, we vary the size of {\wholeM} with three $100\times 100$ low-rank submatrices 
and fix the density at $0.6$.  
In Table~\ref{tab:timesMC}, we see that {\localmc}
decreases the runtime.
This improvement comes from the fact that the completion algorithm becomes more efficient when run on smaller matrices with smaller rank, but also because the rank estimation procedures execute faster.  
\begin{table}[h]
  \centering
\begin{tabular}{c|r|r|r|r|r}
\toprule
$n$    & $800$ & $1000$ & $2000$ & $4000$ & $6000$ \\
\midrule
{\lmafit}     &       $0.76$&  $ 1.34$&  $7.19$& $62.25$& $486.47$   \\ \hline
{\localmc}    &  $ 0.25$ & $ 0.32$&  $2.65$&$ 11.50$&   $69.33$       \\
\bottomrule
\end{tabular}
\caption{The running time of {\lmafit} and {\localmc} in seconds for an $n\times n$ input matrix {\wholeM}. \label{tab:timesMC}}
\end{table}

\spara{Case study with gene-expression data\label{sec:case}:}
To further validate the usefulness of our approach we use a real-world yeast dataset of size
 $2417\times 89$ formed by micro-array expression data~\footnote{The data is described in is described in~\cite{pavlidis2000combining} and can be downloaded at http://mulan.sourceforge.net/datasets-mlc.html}.
 Each row $i$ in the data represents a gene and each column $j$ is an experiment; thus, 
 entry $(i,j)$ is the expression level of gene $i$ in experiment $j$.  
 For each gene, the dataset also provides a phylogenetic 
 profile over 14 groups in the form of a binary vector.

\begin{figure}[h]
\centering
\includegraphics[scale=.5]{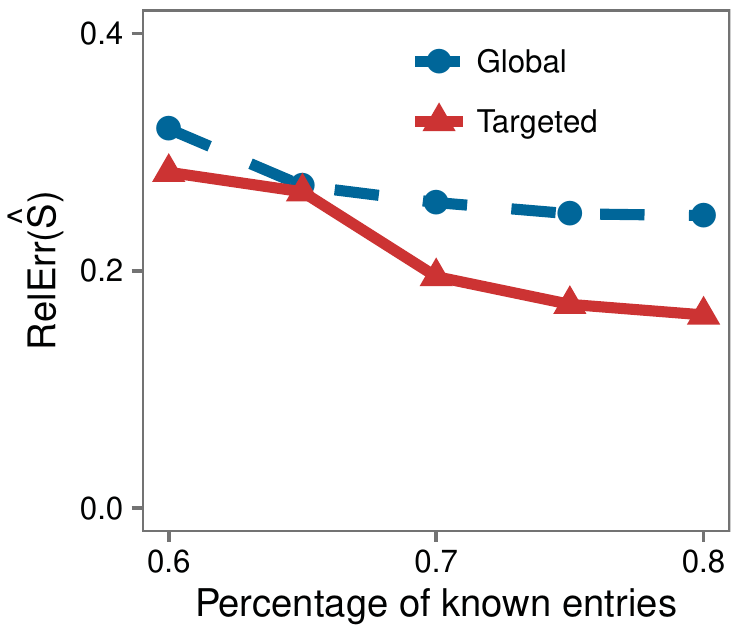}
\includegraphics[scale=.5]{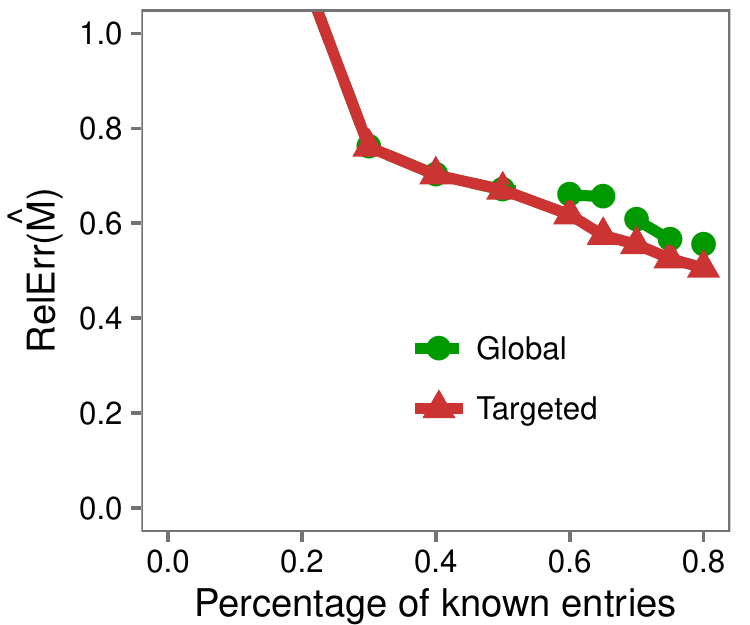}
\caption{~\label{fig:yeastcomp} {\error} of completion on Yeast data.}
\end{figure}

To evaluate {\localmc} on this dataset, we hide a percentage of the known entries just as in the synthetic experiments.   Using a threshold of $\Delta>0.2$ (discussed in Section~\ref{sec:algo}) we evaluate the error over the discovered low-rank submatrix in the estimates produced by the {\lmafit} and {\localmc}. 
Figure~\ref{fig:yeastcomp} shows the {\error} as a function of the percentage of known entries.

Since this matrix data comes with auxiliary attribute information, we go one step further and analyze the submatrix discovered by {\thealgo}.  The identified submatrix {\subM} was $527\times 34$ with $\pi=1.2$.  By comparing the phylogenetic profiles, we found that the genes in {\subM} were similar. Specifically, when we compared the profiles of the genes in {\subM} with the genes not in {\subM}, we found that the average Hamming distance between the profiles of {\subM} was $0.17$ with a median of $0.2$ and a standard deviation of $0.1$, while the average Hamming distance among the rest of the genes was $0.35$ and a median of $0.2$ and a standard deviation of $0.2$. Thus we conclude
that the {\thealgo} helped isolate a subset of genes with similar profile patterns; according to~\cite{pavlidis2000combining}, genes with close profile patterns have similar behavior in biology. 

\spara{Case study with traffic data:} As another real world example we consider a $3000\times 3000$ partially observed Internet traffic matrix.  Each row corresponds to a source Autonomous System (AS), each column corresponds to a destination prefix, and each entry holds the volume of traffic that flowed from an AS to a prefix.  The dataset is only partially observed with $70\%$ of the entries missing.  To mimic the setup of the synthetic experiments, we hid a portion of the $30\%$ known entries and evaluated the accuracy at each step.

\begin{figure}[H]
\centering
\includegraphics[scale=.5]{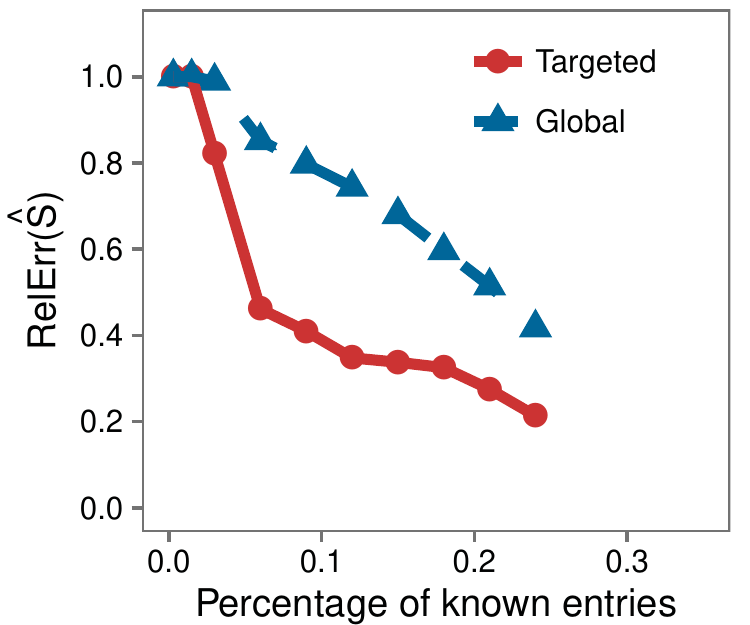}
\includegraphics[scale=.5]{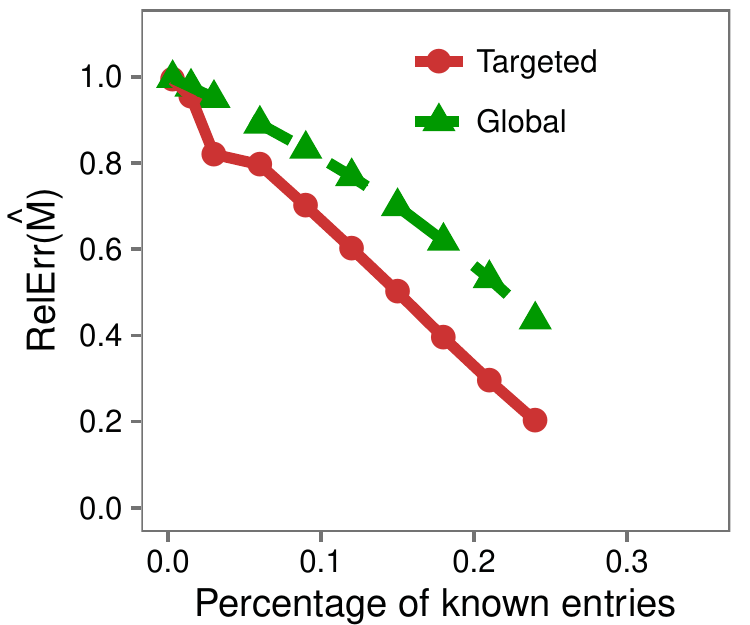}
\caption{~\label{fig:comptraff} {\error} of completion on Traffic data.}
\end{figure}

In Figure~\ref{fig:comptraff} we see the same behavior as on synthetic data, with {\localmc} achieving higher accuracy on the low-rank submatrix.  The effect is less pronounced on the whole matrix {\bM} which we observed to be because the complement {\bT} of the submatrices had high rank, and {\lmafit} is less accurate in this setting.

\label{sec:compexp}

\begin{figure*}
\centering
	\captionsetup[subfigure]{justification=centering}
\begin{subfigure}{0.24\textwidth}
	\centering
\includegraphics[scale=.5]{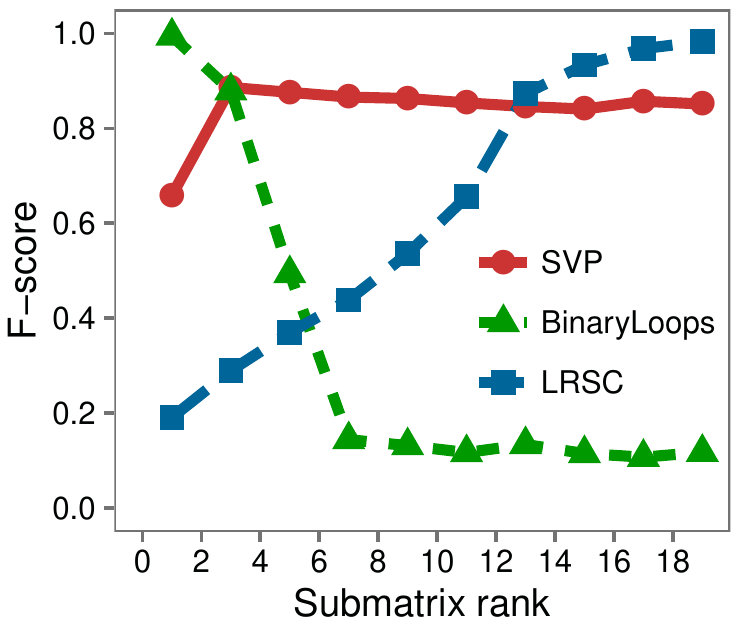}
\caption{Rank $\ell$ of {\subM}.} \label{fig:rank}
\end{subfigure}
\begin{subfigure}{0.24\textwidth}
	\centering
	\includegraphics[scale=.5]{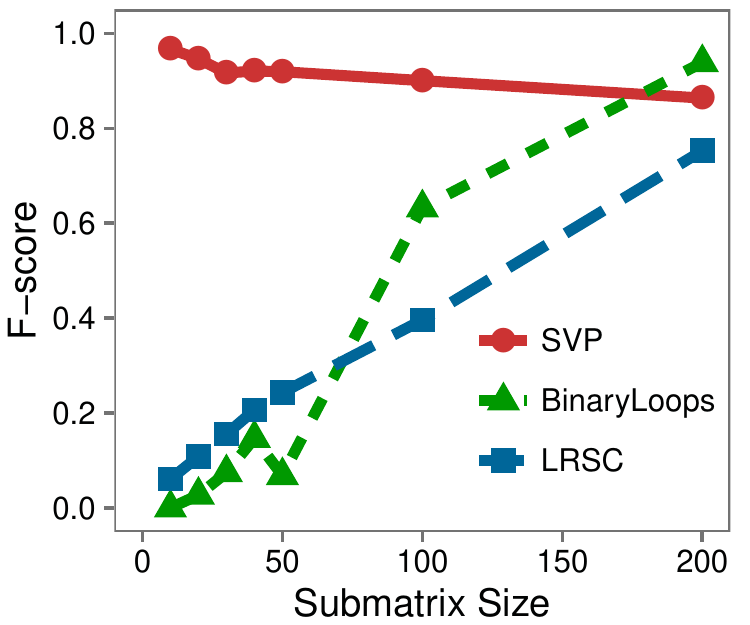}
\caption{Size of {\subM}} \label{fig:size}
\end{subfigure}
\begin{subfigure}{0.24\textwidth}
	\centering
	\includegraphics[scale=.5]{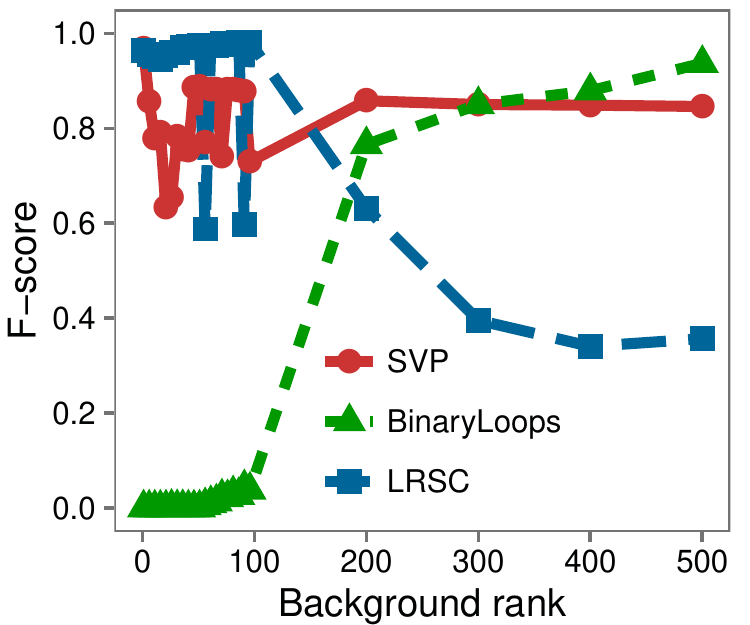}
\caption{Background rank $r$.} \label{fig:backrank}
\end{subfigure}
\begin{subfigure}{0.24\textwidth}
	\centering
	\includegraphics[scale=.5]{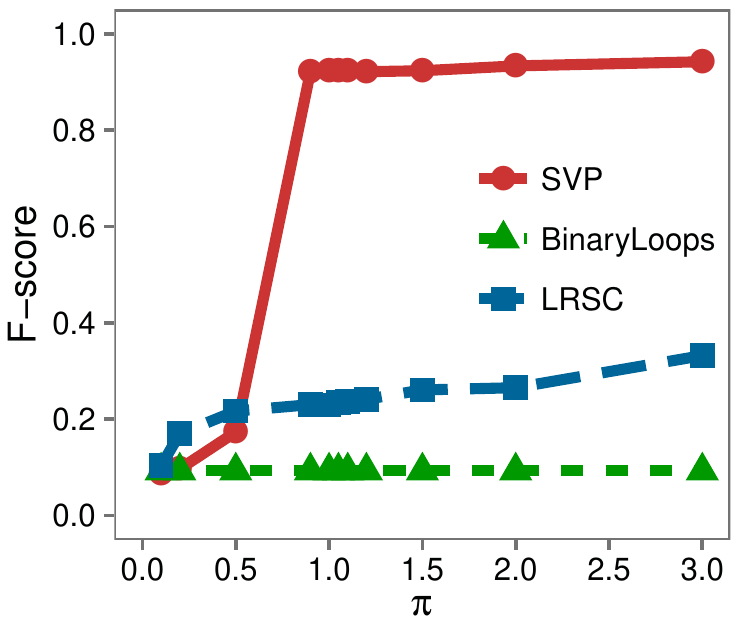}
\caption{$\pi$} \label{fig:pi}
\end{subfigure}
\caption{F-Score achieved by the {\thealgo}, {\tt BinaryLoops}, and {\tt LRSC} algorithms as: the rank $\ell$ of {\subM}, the size of {\subM}, the background rank $r$, or $\pi$ vary.\label{fig:results1}}
\end{figure*}

\subsection{Evaluation of {\thealgo}}
In the previous experiments we have demonstrated the improvement the {\localmc} framework brings to matrix completion.  We now isolate and analyze the algorithmic approach to Step 1 -- the task of finding a low-rank submatrix on \emph{fully known} matrices.
Our results demonstrate that {\thealgo} is effective, efficient, and outperforms 
other heuristics for the same problem for
a large variety of instances.

For experiments we set to rank of the matrix {\wholeM} to $r=1000$ and the rank of the planted submatrix {\subM} to $\ell=5$.
Since the objective of our problem is to find the indices $\rowSub$ and $\colSub$, we evaluate the accuracy of our framework using the combination of precision and recall to the standard F-Score$=2\frac{Pr\times Rcl }{ Pr+Rcl}$. The F-score takes values
in $[0,1]$ and the higher its value the better.

We compare {\thealgo} against the algorithm by Rangan in~\cite{rangan2012simple} (which we call {\tt BinaryLoops})
and algorithms for Subspace Clustering; the approaches are discussed in detail in Section~\ref{sec:related}. For Subspace Clustering we show results for {\tt LRSC} by Vidal and Favaro in~\cite{vidal2014low} since it offers the best balance of accuracy and efficiency; we used the authors' original implementation.   
The baseline algorithms required minor modifications since none of them explicitly output the indices of a submatrix.  For {\tt BinaryLoops} we set $\gamma_{\text{row}}=\gamma_{\text{col}}=0.75$, according to the author's recommendation.  For {\tt LRSC} we set $k=2$ when clustering and select the output $\{\rowSub,\colSub\}$ with the highest accuracy (a user without ground truth would pick the one with lowest empirical rank).  All algorithms were coded in Matlab.

\spara{Varying size and rank:}\label{sec:res}
First we examine a variety of problem instances by comparing the F-Scores of the different algorithms as we vary (a) the rank $\ell$ of {\subM}, (b) the size of {\subM}, (c) the rank $r$ of the background matrix, and (d) $\pi$.  

Figure~\ref{fig:results1} shows that {\thealgo} accurately locates the low-rank submatrix {\subM} for the majority of instances, whether {\subM} is large or small.  In Figure~\ref{fig:pi}, we see that in line with our analysis,  {\thealgo} 
succeeds precisely as the value of $\pi$ grows larger than one.

In contrast to the resilience of {\thealgo} observed in the problem instances, 
{\tt LRSC} and {\tt BinaryLoops} are more sensitive to changes.  {\tt LRSC} performs best when the rank and size of {\subM} are large, 
and the background rank is small. 
On the other hand, {\tt BinaryLoops} has an opposite behavior, performing best on instances where the rank of {\subM} is less than five, and the background rank is large.  
These behaviors are in agreement with the 
analysis and the design of these algorithms in the original papers in which they were introduced.

\begin{figure}[H]  
\centering
\captionsetup[subfigure]{justification=centering}
\begin{subfigure}{0.22\textwidth}
	\centering
 \includegraphics[scale=.5]{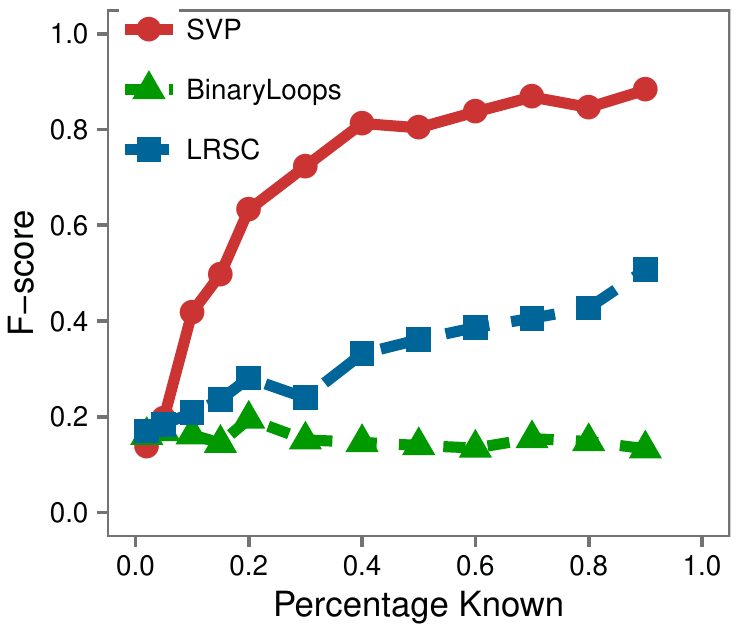}
\caption{\label{fig:incomplete}Incomplete data $\mathbf{M}$.}
\end{subfigure}
\hspace*{\fill} 
\begin{subfigure}{0.22\textwidth}
	\centering
\includegraphics[scale=.5]{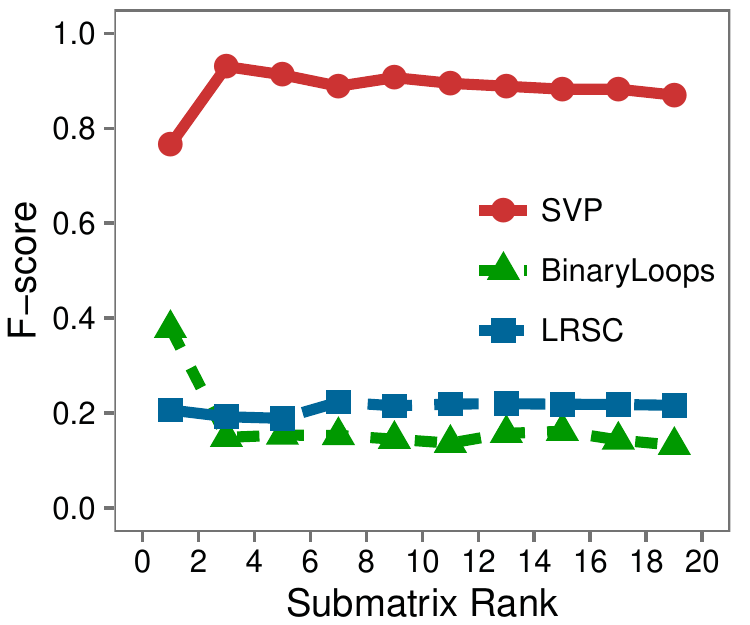}
\caption{Multiple submatrices. \label{fig:many}}
\end{subfigure}
\caption{F-Scores of {\thealgo}, {\tt BinaryLoops}, and {\tt LRSC} on (a) incomplete data as a function of the percentage of known entries, and (b) data with multiple low-rank submatrices as a function of their rank $\ell$.\label{fig:twoplots}}
\end{figure}

\spara{Data with missing entries:}
To compare {\thealgo} to {\tt BinaryLoops}, and {\tt LRSC} on incomplete data, we setup the experiment as described in the setup in Section~\ref{sec:experiments}.
Figure~\ref{fig:incomplete}
shows the F-score of each algorithm as a function of the percentage of known entries.
The results indicate that when the number of known entries is 20\% or above the performance
of {\thealgo} is significantly better than that of {\tt BinaryLoops} and {\tt LRSC}.

\spara{Multiple submatrices:}
Finally 
we test whether the different algorithms are affected by the presence of multiple low-rank submatrices.  We setup the experiment as described in the start of Section~\ref{sec:experiments} and plant a second submatrix $\subM'$ of the same size and rank as {\subM}, with $\pi=1.2$.

Figure~\ref{fig:many} shows the F-Score of each algorithm for the task of finding one submatrix (either {\subM} or $\subM'$) as a function of the ranks $\ell$.  We observe that {\thealgo} significantly outperforms {\tt LRSC} and {\tt BinaryLoops}, and is unaffected by the presence of the second submatrix.  Further, after removal of the first submatrix discovered, {\thealgo} retains the high level of accuracy for subsequent submatrices.

\spara{Running time:}
We compare the running times of {\thealgo}, {\tt BinaryLoops} and {\tt LRSC} as the 
size of the input matrix {\wholeM} increases. The running times of the different algorithms are shown in Table~\ref{tab:times} in seconds, recall all are in Matlab.
\begin{table}[H]
\small
  \centering
	\begin{tabular}{c|r|r|r|r|r}
	\toprule
$n$     & $400$  & $1000$ & $2000$ & $4000$ & $6000$ \\
\midrule
{\thealgo}    &  $0.12 $       &    $0.23$  &  $\mathbf{0.75} $    &  $\mathbf{2.67} $    &   $\mathbf{11.17} $   \\ \hline
{\tt BinaryLoops}     &  $\mathbf{0.02}$&$\mathbf{0.20  }$   &$1.36$   &$8.68$  &   $30.69$       \\ \hline
{\tt LRSC} &  $0.44$   &  $2.19$    &   $9.56$   &   $83.59  $   &  $ 310.87  $   \\
    \bottomrule
    \end{tabular}
		\caption{The running time of {\thealgo}, {\tt BinaryLoops}, and {\tt LRSC} in seconds for an $n\times n$ input matrix {\wholeM}. \label{tab:times}}
\end{table}
We observe that the difference in the running times is more pronounced for large matrices.
For those matrices {\thealgo} is the most efficient one, followed by {\tt BinaryLoops}.
The running time of {\tt LRSC} is $10$ times larger than the running time of {\tt BinaryLoops} and $30$ times larger than that of {\thealgo}. 
Note that the running time of {\thealgo} is dominated by computing the first singular vectors of {\wholeM} and our implementation of {\thealgo} uses the off-the-shelf SVD decomposition of MatLab.
In principle, we could improve this running time by using other SVD speedups and approximations, such as the one proposed in~\cite{drineas06fast}.

\section{Conclusions}\label{sec:conclusions}

The problem of matrix completion is consistently attracting attention and advancement due to its relevance to real-world problems.  Classical approaches to the problem assume that the data comes from a low-rank distribution and fit a single, rank-$r$ model to the whole matrix.  However, many applications implicitly assume that the matrix contains submatrices that are even lower rank.
Despite the popularity of this assumption,
there exist very little work in the matrix completion literature which takes low-rank submatrices into account.

In this work, we propose a {\localmc} framework that explicitly takes into account the presence of low-rank submatrices.  In this framework, the first step is to find low-rank submatrices, and then to apply matrix completion on each component separately.  One of the technical contributions is the development of the {\thealgo} algorithm for find low-rank submatrices in fully and partially known matrices.

Our experiments with real and synthetic data show that {\localmc} increases the accuracy of the completion of matrices containing low-rank submatrices.  Further, the experiments demonstrate that {\thealgo} is an efficient and effective algorithm for extracting low-rank submatrices.

\bibliographystyle{abbrv}
\bibliography{local}

\begin{thebibliography}{10}

\bibitem{aggrawal16recommender}
C.~C. Aggarwal.
\newblock {\em {Recommender Systems}}.
\newblock Springer, 2016.

\bibitem{alon1998finding}
N.~Alon, M.~Krivelevich, and B.~Sudakov.
\newblock Finding a large hidden clique in a random graph.
\newblock {\em Random Structures and Algorithms}, 13(3-4):457--466, 1998.

\bibitem{incpack}
C.~Baker.
\newblock Incremental svd package.
\newblock \url{http://www.math.fsu.edu/~cbaker/IncPACK/}, 2012.

\bibitem{cai2010singular}
J.-F. Cai, E.~J. Cand{\`e}s, and Z.~Shen.
\newblock A singular value thresholding algorithm for matrix completion.
\newblock {\em SIAM Journal on Optimization}, 20(4):1956--1982, 2010.

\bibitem{candes12exact}
E.~J. Cand{\`e}s and B.~Recht.
\newblock Exact matrix completion via convex optimization.
\newblock {\em Commun. ACM}, 2012.

\bibitem{chen14coherent}
Y.~Chen, S.~Bhojanapalli, S.~Sanghavi, and R.~Ward.
\newblock Coherent matrix completion.
\newblock In {\em International Conference on Machine Learning (ICML)}, 2014.

\bibitem{davenportoverview}
M.~Davenport and J.~Romberg.
\newblock An overview of low-rank matrix recovery from incomplete observations.

\bibitem{drineas06fast}
P.~Drineas, R.~Kannan, and M.~W. Mahoney.
\newblock Fast monte carlo algorithms for matrices ii: Computing a low-rank
  approximation to a matrix.
\newblock {\em SIAM Journal on Computing}, 36(1):158--183, 2006.

\bibitem{elhamifar2009sparse}
E.~Elhamifar and R.~Vidal.
\newblock Sparse subspace clustering.
\newblock In {\em Computer Vision and Pattern Recognition, 2009. CVPR 2009.
  IEEE Conference on}, pages 2790--2797. IEEE, 2009.

\bibitem{foygel2011concentration}
R.~Foygel and N.~Srebro.
\newblock Concentration-based guarantees for low-rank matrix reconstruction.
\newblock In {\em COLT}, pages 315--340, 2011.

\bibitem{funksvd}
S.~Funk.
\newblock Incremental svd for the netflix prize.
\newblock \url{http://sifter.org/~simon/journal/20061211.html}, 2006.

\bibitem{gionis2007clustering}
A.~Gionis, H.~Mannila, and P.~Tsaparas.
\newblock Clustering aggregation.
\newblock {\em ACM Transactions on Knowledge Discovery from Data (TKDD)},
  1(1):4, 2007.

\bibitem{goldberg1992using}
D.~Goldberg, D.~Nichols, B.~M. Oki, and D.~Terry.
\newblock Using collaborative filtering to weave an information tapestry.
\newblock {\em Communications of the ACM}, 35(12):61--70, 1992.

\bibitem{herlocker04evaluating}
J.~L. Herlocker, J.~A. Konstan, L.~G. Terveen, and J.~T. Riedl.
\newblock Evaluating collaborative filtering recommender systems.
\newblock {\em ACM Trans. Inf. Syst.}, 22(1):5--53, 2004.

\bibitem{keshavan2009matrix}
R.~H. Keshavan, S.~Oh, and A.~Montanari.
\newblock Matrix completion from a few entries.
\newblock In {\em 2009 IEEE International Symposium on Information Theory},
  pages 324--328. IEEE, 2009.

\bibitem{lee2013local}
J.~Lee, S.~Kim, G.~Lebanon, and Y.~Singer.
\newblock Local low-rank matrix approximation.
\newblock In {\em Proceedings of The 30th International Conference on Machine
  Learning}, pages 82--90, 2013.

\bibitem{meka2009matrix}
R.~Meka, P.~Jain, and I.~S. Dhillon.
\newblock Matrix completion from power-law distributed samples.
\newblock In {\em Advances in neural information processing systems}, pages
  1258--1266, 2009.

\bibitem{pavlidis2000combining}
P.~Pavlidis and W.~N. Grundy.
\newblock Combining microarray expression data and phylogenetic profiles to
  learn gene functional categories using support vector machines.
\newblock 2000.

\bibitem{rangan2012simple}
A.~V. Rangan.
\newblock A simple filter for detecting low-rank submatrices.
\newblock {\em Journal of Computational Physics}, 231(7):2682--2690, 2012.

\bibitem{rennie2005fast}
J.~D. Rennie and N.~Srebro.
\newblock Fast maximum margin matrix factorization for collaborative
  prediction.
\newblock In {\em Proceedings of the 22nd international conference on Machine
  learning}, pages 713--719. ACM, 2005.

\bibitem{ruchansky2015matrix}
N.~Ruchansky, M.~Crovella, and E.~Terzi.
\newblock Matrix completion with queries.
\newblock In {\em Proceedings of the 21th ACM SIGKDD International Conference
  on Knowledge Discovery and Data Mining}, pages 1025--1034. ACM, 2015.

\bibitem{sarwar01itembased}
B.~Sarwar, G.~Karypis, J.~Konstan, and J.~Riedl.
\newblock Item-based collaborative filtering recommendation algorithms.
\newblock In {\em International Conference on World Wide Web}, WWW, pages
  285--295. ACM, 2001.

\bibitem{vidal2014low}
R.~Vidal and P.~Favaro.
\newblock Low rank subspace clustering (lrsc).
\newblock {\em Pattern Recognition Letters}, 43:47--61, 2014.

\bibitem{wen2012solving}
Z.~Wen, W.~Yin, and Y.~Zhang.
\newblock Solving a low-rank factorization model for matrix completion by a
  nonlinear successive over-relaxation algorithm.
\newblock {\em Mathematical Programming Computation}, 4(4):333--361, 2012.

\end{thebibliography}

\end{document}